\documentclass[letterpaper]{article} 
\usepackage[]{aaai23}  
\usepackage{times}  
\usepackage{helvet}  
\usepackage{courier}  
\usepackage[hyphens]{url}  
\usepackage{graphicx} 
\usepackage{natbib}  
\usepackage{caption} 
\usepackage{tabularx}
\usepackage{subcaption}
\usepackage{multirow}

\usepackage{algorithm}
\usepackage{algorithmic}

\usepackage{newfloat}
\usepackage{listings}
\usepackage{amsmath,amsfonts}
\usepackage{amssymb} 
\DeclareCaptionStyle{ruled}{labelfont=normalfont,labelsep=colon,strut=off} 
\floatstyle{ruled}
\newfloat{listing}{tb}{lst}{}
\floatname{listing}{Listing}

\title{Imputing Knowledge Tracing Data with Subject-Based  Training via LSTM Variational Autoencoders Frameworks}

\author {
    Jia Tracy Shen\hspace{0.5in}
    Dongwon Lee 
}
\affiliations {
    The Pennsylvania State University, USA\\
    \{jqs5443, dongwon\}@psu.edu
    }
\usepackage{bibentry}

\begin{document}

\maketitle

\begin{abstract}
The issue of missing data poses a great challenge on boosting performance and application of deep learning models in the {\em Knowledge Tracing} (KT) problem. However, there has been the lack of understanding on the issue in the literature.
In this work, to address this challenge, we adopt a subject-based training method to split and impute data by student IDs instead of row number splitting which we call non-subject based training. The benefit of subject-based training can retain the complete sequence for each student and hence achieve efficient training. Further, we leverage two existing deep generative frameworks, namely variational Autoencoders (VAE) and Longitudinal Variational Autoencoders (LVAE) frameworks and build LSTM kernels into them to form LSTM-VAE and LSTM LVAE (noted as VAE and LVAE for simplicity) models to generate quality data. In LVAE, a Gaussian Process (GP) model is trained to disentangle the correlation between the subject (i.e., student) descriptor information (e.g., age, gender) and the latent space. The paper finally compare the model performance between training the original data and training the data imputed with generated data from non-subject based model VAE-NS and subject-based training models (i.e., VAE and LVAE). We demonstrate that the generated data from LSTM-VAE and LSTM-LVAE can boost the original model performance by about 50\%. Moreover, the original model just needs 10\% more student data to surpass the original performance if the prediction model is small and 50\% more data if the prediction model is large with our proposed frameworks. 
\end{abstract}

\section{Introduction}

\begin{figure}
    \centering
    \includegraphics[width=\columnwidth]{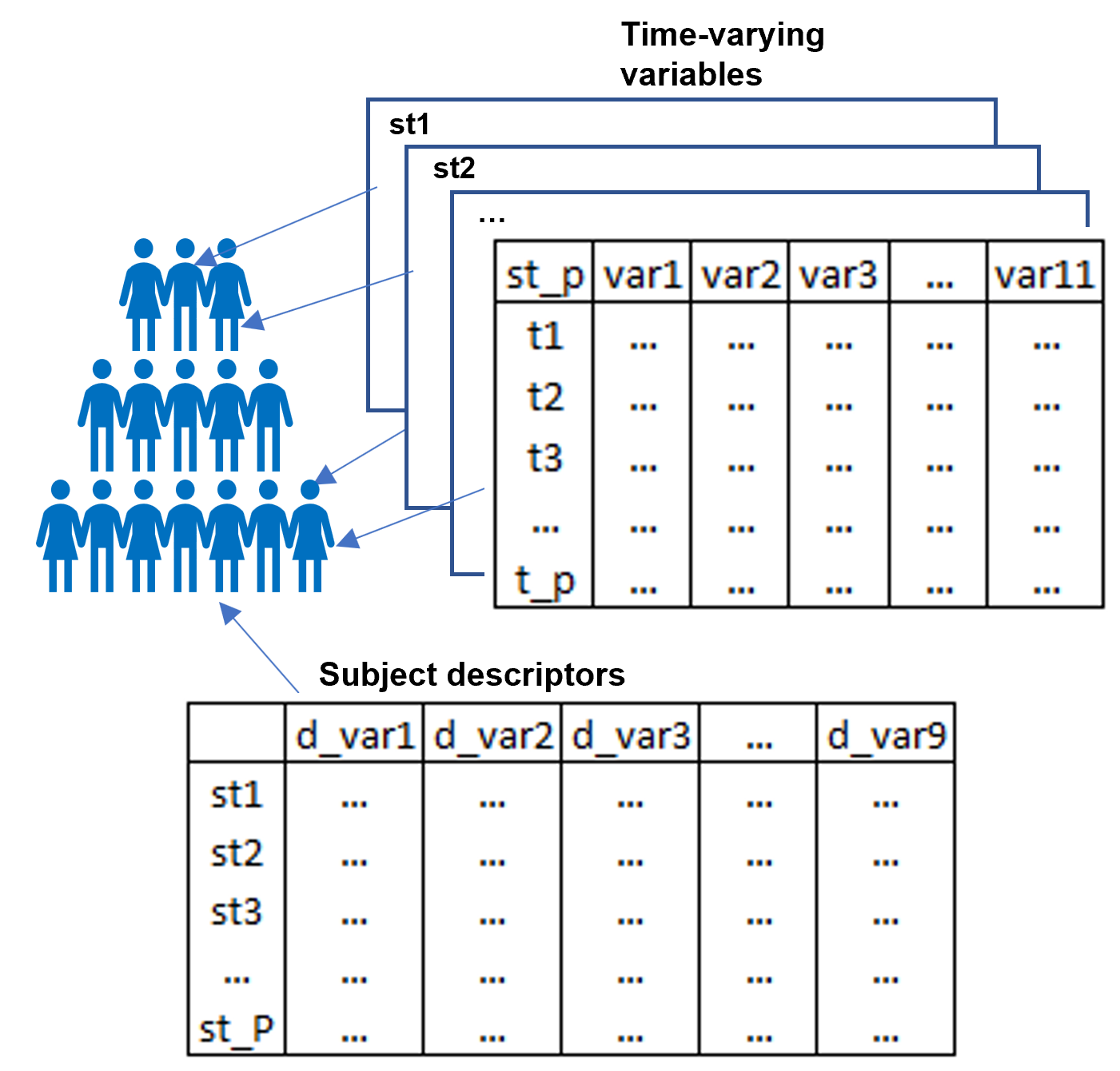}
    \caption{An Illustration of the Longitudinal (student) Data in KT Field.  `p': student p. `P': total \# of students.}
    \label{fig:subject_data}
\end{figure}

Knowledge tracing (KT) as a student modeling technique has been widely used to predict and trace students' knowledge state during their learning processes. In recent years, with the huge success that deep learning has brought to the field, there are many KT algorithms that can predict individuals' knowledge state to a decent extent. However, the \textit{sparseness} of students' exercise data represented by \textit{missing values} still limits the models' performance and application \cite{Swamy2018DeepProgression}. About half of the existing publications use public data sets \cite{Dai2021KnowledgeTechnologies}, which can not be available for huge amount due to administration cost. Researchers could opt for other private data sets that however may not even have the sizable volume as the public data sets. Besides, many deep learning algorithms including the state-of-art (SOTA) KT algorithms need huge and diverse amount of training data to obtain decent performances. On the other hand, it is unavoidable to see the missing values in KT data because of two reasons: (i) data is missed completely at random (MCAR) where the probability of missing data is independent on its own value and on other observable values \cite{RoderickJ.A.Little2002StatisticalData}. For example, due to COVID, we have many students missing exams; (ii) the data is missed not at random (MNAR), which indicates the reason for a missing value can depend on other variables but also on the value that is missing. For example, if a student performs poorly on the English subject and often miss exams in other subjects, his missed records in English quizzes can be attributed to other known reasons. 
Moreover, KT data is a type of longitudinal data, all collected repeatedly over time for each \textit{subject} (ie., student). Such data contains both dependent and independent variables. For example, the dependent variables in KT data can comprise time-varying measurements per \textit{subject} (e.g., response correctness, time taken per question), whereas independent variables are time-invariant \textit{subject descriptors} (e.g., grade, gender, gifted or not) (see the illustration in Figure \ref{fig:subject_data}). Analyzing such data is challenging as it often includes high-dimensional time [in]variant variables with missing values. Despite that missing data in KT field is ubiquitous and poses challenges on achieving better model results, there are very few studies researching on effective approaches to tackle the missing data issue in KT field. Our work is one of the few studies to address such challenge. 

To that end, we suppose a deep generative model such as Variational Autoencoders (VAE) \cite{Kingma2019AnAutoencoders} could effectively generate data for the missing values because of its superiority over other generative models (e.g., Generative Adversarial Networks) in time series data generation \cite{Le2020GRACE:Prediction,Fahrmann2022LightweightSystems}. Furthermore, given the challenge arisen from the longitudinal KT data, we make two hypotheses: (i) a training style that can reflect the subject longitudinal nature could entail more effective training; (ii) the information from subject descriptors could potentially represent the latent space better and help improve the quality on the data generation. To validate hypothesis (i), we develop a subject-based training style where we split and impute data by student IDs to reflect the longitudinal nature of the subjects. The benefit of doing so is to maintain the complete sequence for each student whereas splitting by row number could separate the individual sequence and entail inefficient training. Thus, applying subject-based training on top of VAE framework could potentially address the challenge. To validate hypothesis (ii), we leverage a module from the existing Longitudinal VAE (LVAE) \cite{Ramchandran2021LongitudinalAutoencoder} framework called additive multi-output Gaussian Process (GP) prior that can extrapolate the correlation between time-invariant subject descriptors and the latent space to enhance the latent variable learning. Given the longitudinal nature of the LVAE framework, a subject-based training can be naturally applied to LVAE to boost data generation quality. Furthermore, we build LSTM kernels to both VAE and LVAE frameworks because LSTMs are good at extrapolating the temporal relationship from multi-variate time series data \cite{Pearlmutter1989LearningNetworks,Giles1994NeuralApplications}. With the generated data from the proposed frameworks, we will be able to impute them back for retraining and evaluate the effectiveness of the imputed data on boosting the original model performance. Besides, we are also interested to discover how robust our generated data can be on boosting the original model performance, e.g., by only applying a fraction of the generated data. 
 
Thus, this work attempts to make the following contributions:

\begin{itemize}
    \item Overcoming the issue of the missing KT data, we conduct \textit{subject-based} training on KT data via LSTM-VAE framework
    \item Leveraging the additive GP prior module from LVAE, we form a LSTM-LVAE framework to showcase the superiority of training additional \textit{subject descriptors} for better latent space representation
    \item We demonstrate the robustness of only using a fraction of the generated data to boost the original model performance
\end{itemize}  

\section{Method}

We propose two deep generative frameworks: LSTM-VAE and LSTM-LVAE. The both frameworks use subject-based training. We explain the details as follows.

\subsection{Problem setting}
According to \citeauthor{Ramchandran2021LongitudinalAutoencoder} \citeyear{Ramchandran2021LongitudinalAutoencoder},
let $D$ be the dimensionality of the observed data, $P$ be the number of unique students, $n_{p}$ be the total number of longitudinal samples from student $p$, and $N=\sum_{p=1}^P n_{p}$ be the total number of samples. Therefore, the longitudinal samples for student $p$ are denoted as $Y_{p}=[y^p_{1},...,y^p_{n_{p}}]^T$, where each sample $y^p_{t} \in \mathcal{Y}$ and $\mathcal{Y}=\mathbb{R}^D$. The subject descriptors for students are represented as $X_{p}=[x^p_{1},...,x^p_{n_{p}}]^T$, where $x^p_{t} \in \mathcal{X}$ and $\mathcal{X}=\mathbb{R}^Q$, $Q$ be the number of descriptors. The latent space is then denoted as $\mathcal{Z}=\mathbb{R}^L$ and a latent embedding for all $N$ samples as $Z=[z_{1},...,z_{N}]^T \in \mathbb{R}^{N \times L}$ with $L$ being the number of latent dimensions. To generate data, a joint generative model is then parameterized by $w=\{\psi,\theta\}$ as $p_{w} (y,z)=p_{\psi} (y|z) p_{\theta} (z)$. Therefore, if the latent variable $z$ is known, it will be easy to infer $y$ and hence generate the desired data. 
\begin{figure*}
    \centering
    \includegraphics[width=\linewidth]{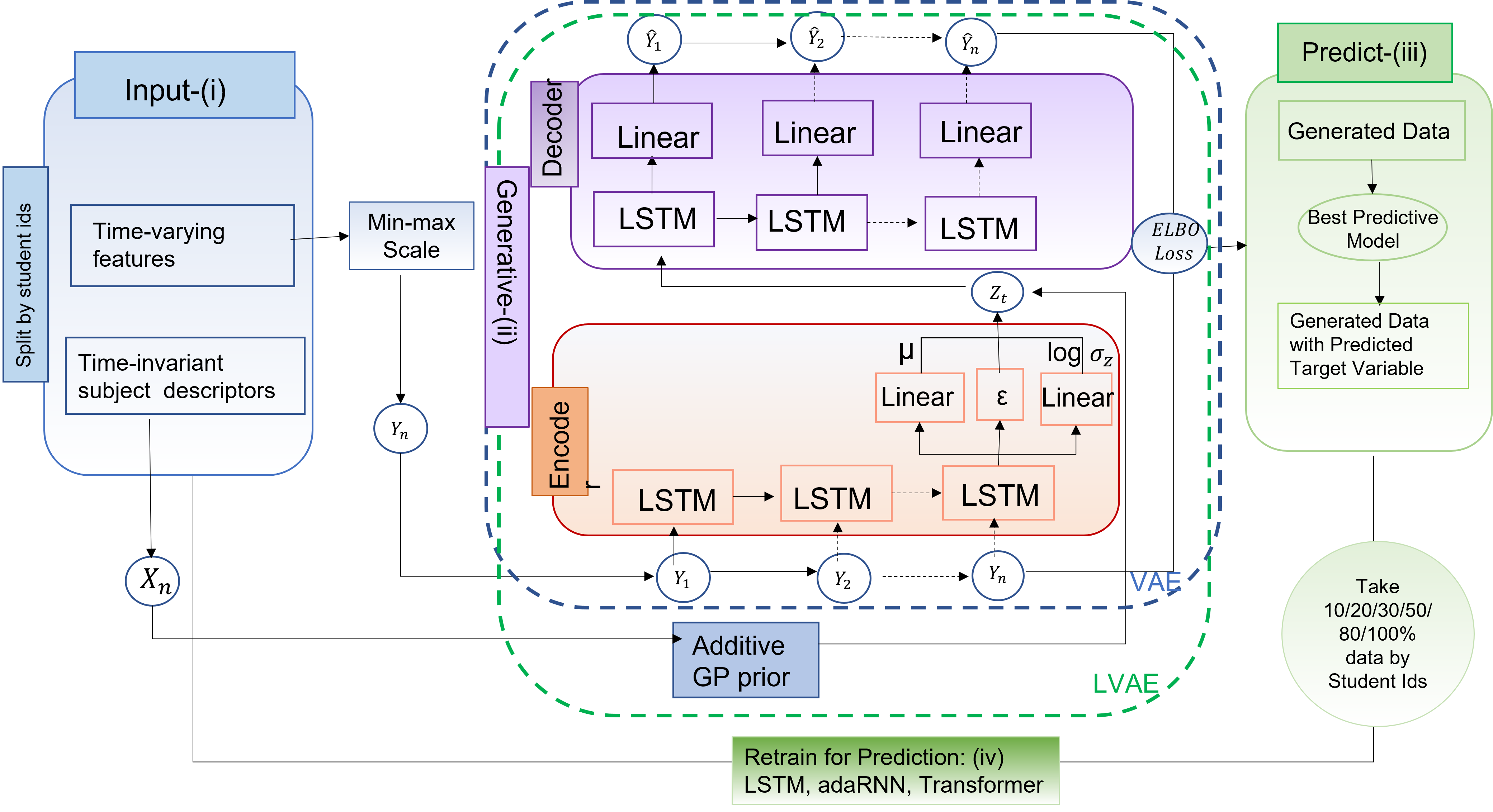}
    \caption{Overview of the methodology proposed in this work.}
    \label{fig:methods}
\end{figure*}

\subsection{VAE and LVAE}
To infer the latent variable $z$ given $y$, the posterior distribution is $p_{w} (z|y)=p_{\psi} (y|z) p_{\theta} (z)/p_{w} (y)$ and is generally intractable due to the marginalization over the latent space $p_{w} (y)=\int p_{\psi}(y|z) p_{\theta} (z) dz $. Therefore, Variational Auto-Encoder \cite{Kingma2019AnAutoencoders} introduced the approximated version posterior, noted as $q_{\phi} (z|y)$ instead of the true posterior $p_{w} (z|y)$ and fit the approximate inference model by maximizing the Evidence Lower Bound (ELBO) of the marginal log-likelihood w.r.t. $\phi$:\[\log p_{w} (Y)\ge \mathcal{L} (\phi,\psi,\theta; Y)\]\[ \triangleq \mathbb{E}_{q_{\phi}} [\log p_{\psi} (Y|Z)] -D_{KL} (q_{\phi} (Z|Y)||p_{\theta}(Z))\rightarrow \underset{\phi}{\max} ,
\] where $\mathbb{E}_{q_{\phi}} [\log p_{\psi} (Y|Z)]$ is a reconstruction error, measuring the difference between the input and the encoded-decoded data. and $D_{KL}$ denotes the Kullback-Leibler Divergence (KLD), measuring the divergence between $q_{\phi} (Z|Y)$ and $p_{\theta} (Z)$. In practice, we minimize the negative ELBO: $ D_{KL} (q_{\phi} (Z|Y)||p_{\theta} (Z)-\mathbb{E}_{q_{\phi}} [\log p_{\psi} (Y|Z)]$, where all the parameters are learned simultaneously together: $\mathcal{L} (\phi,\psi,\theta; Y) \rightarrow \min_{\phi,\psi,\theta}$. 

When facing the longitudinal data, \citeauthor{Ramchandran2021LongitudinalAutoencoder} hypothesize $z$ has relationship with both $Y$ and $X$ and formulate the generative model as 
\[p_{w} (Y|X)=\int_{Z} p_{\psi} (Y|Z,X) p_{\theta} (Z|X) dZ\] 
\[=\int_{Z} \prod p_{\psi} (y_{n}|z_{n}) p_{\theta} (Z|X) dZ,\]where $p_{\psi} (y_{n}|z_{n})$ is normally distributed probabilistic decoder and $p_{\theta} (Z|X)$ is defined by the multi-output additive GP prior that regulates the joint structure of $Z$ with descriptors $X$. The Additive GP is a Gaussian process prior as $f(x) \sim GP( \mu (x), K(x,x^\prime|\theta))$, where $\mu (x) \in \mathbb{R_{L}} $ is the mean (assumed as 0) and $K(x,x^\prime|\theta)$ is a matrix-valued positive definite cross-covariance function (CCF). Based on the practice of \citeauthor {Cheng2019AnData}, LVAE constructs the additive GP components with squared exponential CFs (from continuous variables), categorical CFs (from categorical covariates),the interaction CFs (the product of the categorical and squared exponential CFs) and the product of the squared exponential CFs and the binary CFs. Finally, the ELBO function changes to the following after factoring the descriptors $X$:\[\log p_{w} (Y|X)\ge \mathcal{L} (\phi,\psi,\theta; Y, X)\]
\[\triangleq \mathbb{E}_{q_{\phi}} [\log p_{\psi} (Y|Z)] -D_{KL} (q_{\phi} (Z|Y)||p_{\theta} (Z|X)\rightarrow \underset{\phi}{\max}.\] LVAE differentiates from VAE in that it hypothesizes there exists a relationship between $\mathcal{X}$ and the latent space $\mathcal{Z}$ and uses an additive multi-output Gaussian Prior to extract that relationship.

\subsection{Generative Frameworks}
Based on above solutions, two generative frameworks are developed (see in Figure \ref{fig:methods}). It has 4 phases: (i) input phase that pre-processes data; (ii) generative phase where data gets generated via the generative model framework; (iii) prediction phase where we predict target variable for the generated data; (iv) retraining phase, where we combine the original data and generated data to retrain for donwstream prediction task.

From the figure, after input phase (i), we see that the data gets separated into two sets: (a) time-varying data (noted as $Y=\{y_{1},...,y_{n}\}$); (b) time-invariant subject descriptors (noted as $X=\{x_{1},...,x_{n}\}$). The time-varying data $y_{n}$ goes through a min-max scaler, a typical time series data normalization method \cite{Yu2021UsingEnvironment}, and enters the LSTM encoder to generate $\mu$ and log $\sigma_{z}$ for the latent distribution ${Z_{t}}$. The time-invariant subject descriptors $X_{n}$ on the other hand are only fed into the Additive GP prior module to train for the approximated GP prior with its output merging into the latent space ${Z_{t}}$. Next, the decoder samples on the latent distribution and reconstructs data {$\hat{Y}_{n}$}, namely encoded-decoded data, based on the latent features from ${Z_{t}}$. Here, we name the generative framework that only includes the encoder and decoder as LSTM-VAE and the framework that includes encoder, decoder and the additive GP prior module as LSTM-LVAE. We omit LSTM prefix for simplicity. After that, we compare $Y_{n}$ to $\hat{Y}_{n}$ for evaluation via ELBO. VAE assigns the equal weight for both reconstruction and KLD errors whereas LVAE assigns a weight to KLD to regularize further. Once we have good generation quality, we generate data on the missing data which has all the subject descriptor information but missing on all the time-varying features. 

Before entering phase (iii), we conduct initial prediction on the original data via models that work well with multi-variate time series data: LSTM, adaRNN and Transformer. Similar to LSTM, adaRNN (i.e., adaptive RNN) \cite{Du2021AdaRNN:Series} is a recurrent neural network but based on the Gated Recurrent Unit that comprises two gates (i.e., reset gate and update gate). It usually trains faster than LSTM and easy to modify and works better if the sequence is not too long. Because some KT data could present non-sequential characteristics, we include the adapted version of the original Transformer model \cite{Vaswani2017AttentionNeed}, whose attention mechanism and positional embedding are great for non-sequential data. To evaluate these models, we use Root Mean Square Error (RMSE) as our target variable is continuous (i.e., score rate, the possible score obtained per question divides the total scores obtained per student). After the initial round of prediction is performed, we conduct phase (iii) by selecting the best predicting model to predict the target variable for the generated data from phase (ii). In phase (iv), we impute the fraction of 10/20/30/50/80/100\% of the generated data (with target variable) back to the original data and retrain for the downstream predictive task. 

\subsection{Subject-based Training}
\begin{figure}
    \centering
    \includegraphics[width=\linewidth]{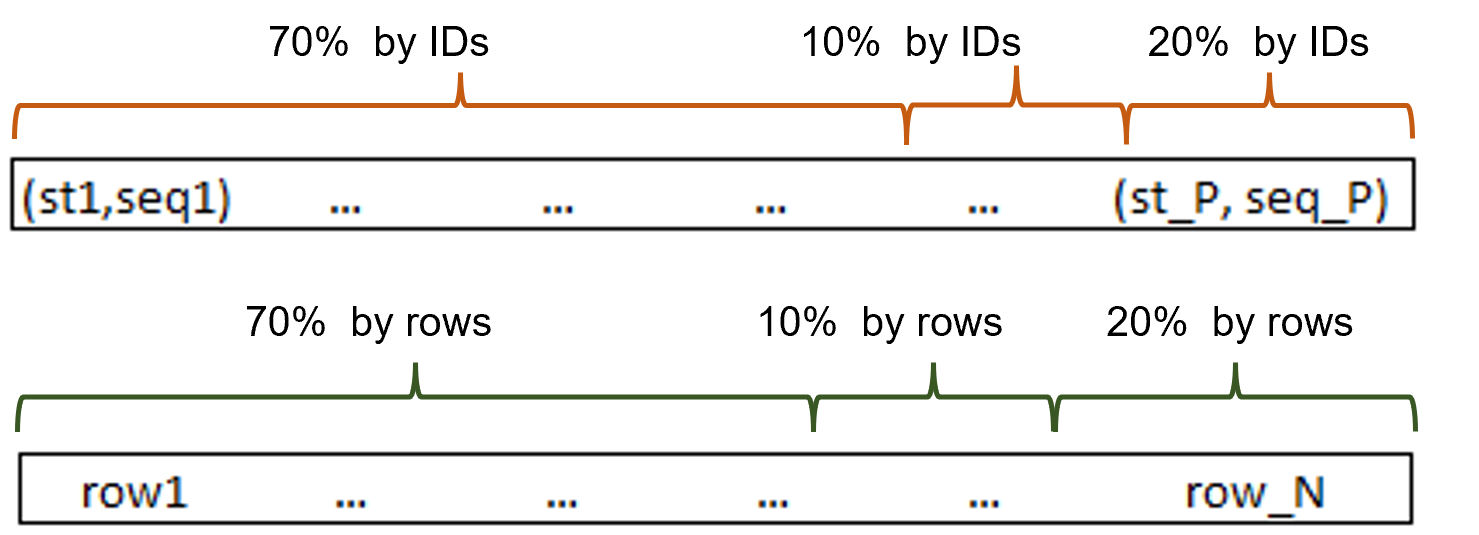}
    \caption{An Illustration of Split by IDs vs. Row Number}
    \label{fig:data_process}
\end{figure}

Besides the generative frameworks, this work also takes a new training strategy, that is, subject-based training. We refer subject-based training to a style where data are split and imputed back by student IDs instead of row number. We call the training using row-number splitting as non-subject based training. For example, in subject-based setting, 70\% of student IDs are extracted as training data and 10\% student IDs are extracted as validation data whereas in non-subject based setting, 70\% of total rows are extracted as training data and 10\% total rows are extracted as validation data) (see the illustration in Figure \ref{fig:data_process}). We see the split points by IDs are not the same as splitting by row number. It indicates there is chance that the sequence of certain students will be cut into two pieces, leaving them into two different sets (e.g, val and test). If we split the data by student IDs, we can impute the generated data back to the original data via IDs and keep the learning sequence relevant and complete for each student. If we opt for row-number splitting, the student's original sequence will be interfered and not be trained appropriately.   

\section{Experiments}
In this section, we carry out two major experiments. The first experiment is to generate the data and impute back to the original data for retraining. It has three steps: (a) generate knowledge tracing data by utilizing VAE and LVAE; (b) predict the target variable for the generated data using the best model obtained from the KT prediction task; (c) merge the generated data with the original data to retrain for model performance. The second experiment is to validate the robustness of imputed data on boosting the original model performance. More specifically, we add a fraction of the generated data in the cadence of 10\%, 20\%, 30\%, 50\%, 80\%, 100\% during the retraining phase to examine the boosting effect. 

\begin{figure}
    \centering
    \includegraphics[width=\linewidth]{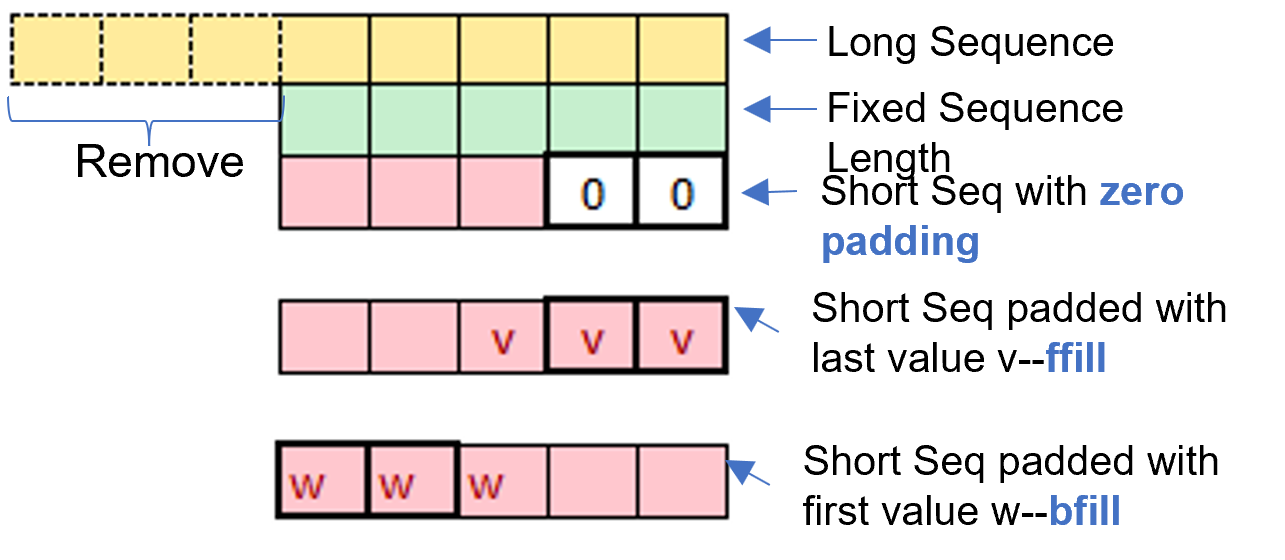}\\
     
    \caption{An illustration of KT data sequence aligning process and 3 Padding Strategy.}
    \label{fig:3_seq}
\end{figure}

\subsection{Data sets}

To achieve above, we need to apply our model onto the data sets that have subject descriptors so that we can use LSTM-LVAE model to generate missing data. Unfortunately, the public data sets (e.g., ASSISTment datasets, Junyi, STATICS, EdNet, etc) in KT field do not contain subject descriptor information such as the student's grade level, gifted or not. This is also why the renowned deep learning models such as DKT, DKVMN, NPA, SAINTS \cite{Minn2018DeepTracing,Zhang2017DynamicTracing, Pandey2019ATracing, Shin2021SAINT+:Prediction} are not included in the chapter because most of these models are generated for single variable KT data or take data feature as hyper-parameters. Thus, we use the two private multivariate KT data sets from K12.com platform (an online K-Grade 12 education platform). They are : (1) Grade 10 geometry course (noted as Geom) quiz answering data set with average sequence length of 150 time steps; (2) Grade 11 algebra II (noted as Alg2) quiz answering data set with average sequence length of 150 time steps. Each data set contains 11 temporal features (i.e. sequence number, assessment duration, attempts per question, total attempts, question difficulty, item difficulty, standard difficulty, question reference, item reference, standard id, question type) from July 2017 to June 2019 and 7 subject descriptors that define the student profiles (i.e., school ID, special ED, student id, free reduced lunch, gifted\_talented, grade level, score rate). The Geom data set contains 3,265 total students with 412,397 observed instances whereas the Alg2 data has 2,110 total students with 277,548 observed instances.

\subsection{Identify Missing Values} 
In practice, it is hard to identify the missing steps each student has because their learning experience varies. Thus, we develop a regime where we first find all the quiz times of a school where the student is located and then fill up the missing times by comparing to the school's full quiz taking schedule. For example, if school A has 100 quiz times but student A only has 60 records, we fill out the remaining 40 quiz time steps based on the event time variable. This approach is a bit rigorous, assuming all the students are required to test for the same number of quizzes if they are in the same school and skipping any quiz is considered as a missing step. In reality, there might be scenarios where students are allowed to skip, which is complex to study and hence we use this approach as it is straightforward.  With that, we are able to retrieve the missing time steps before and after the current temporal steps for all the students. As the students are known, this missing data has all the subject descriptor information. 

\subsection{Data Processing}
To conduct the training for generation, we split the data by 0.5/0.1/0.2/0.2 for train/val/test/generate and 0.7/0.1/0.2 for train/val/test during downstream prediction (see in Table \ref{tab:data}). Note that the generation set with a ratio of 0.2 is used to evaluate the quality of generation whereas the generated set we use to impute back to the original data is generated from missing data. Based on the above missing data identification regime, we are able to identify 3,233 out of 3,265 total students who have a total of 799,408 missed instances from the Geom course and 2,057 out of 2,110 students who have a total of 516,884 missed instances from the Alg2 course (see in Table \ref{tab:data}). Because all the ratios are applied to both subject and non-subject based training, the generated data from missing values will be imputed back to the original data via IDs in the subject-based training and via row number in the non-subject based training in the splits of train/val/test. Both training styles align data to a fixed sequence length which is due to the model input requirement of 3D dimensions (i.e., batch size * sequence length * number of dimensions). This also aligns with the typical data processing technique for KT model training \cite{Pardos2011KT-IDEM:Model, Lee2019CreatingAssess, Pandey2019ATracing}. If the actual student learning sequence is longer than the fixed sequence length, we cut the part where it exceeds. If the sequence is shorter than the fixed sequence, we pad it (see in Figure \ref{fig:3_seq}). We use three padding strategies to find an optimal model performance: (a) zero paddding; (b) ffill; (c) bfill. Ffill pads forward with the last value `v' whereas bfill pads backward with the first value `w'  (see in Figure \ref{fig:3_seq}). Bfill in practice assumes that a student gets the same quiz result in his missed quiz as his first quiz result whereas ffill assumes that a student gets the same quiz result in his missed quiz as his last quiz result. Zero-padding just simply assumes that a student gets zero in his missing quiz.

\begin{table}
  \caption{Data Statistics for Geom and Alg2 Data. * is Downstream Task Split}
  \label{tab:data}
  \resizebox{\columnwidth}{!}{
  \begin{tabular}{|c|c|c|c|c|}
    \hline
   \multirow{2}{*}{Split Part (Ratio)}&\multicolumn{2}{c|}{Geometry (Geom) }&\multicolumn{2}{c|}{Algebra II (Alg2)} \\ 
    \cline{2-5}
     & \# Student& \# of Rows &\# Student& \# of Rows\\
    \hline
    Train (0.5) &1,633 &215,632 &1,055 & 137,409 \\
 
    Validate (0.1)&326 & 42,259& 211& 30,652 \\
   
    Test (0.2) & 653& 82,707&422 & 60,709 \\
   
    Generation (0.2) &653 &71,799 &422 & 48,778 \\
    Data Total &3,265&412,397&2,110&277,548\\
    \hline

    Train* (0.7) &2,286 &287,431&1,477 & 186,187 \\
 
    Validate* (0.1) & 326&42,259 &211 & 30,652 \\
   
    Test* (0.2) &653 & 71,799&422 &60,709  \\
    Data Total* &3,265&412,397&2,110&277,548\\
    \hline

    Missing Train (0.7) &2,256 &559,586&1,440 & 361,819 \\
 
    Missing Validate (0.1) & 322&79,941 &206 & 51,688 \\
   
    Missing Test (0.2) &645 & 159,882&411 &103,377  \\
      Missing Data Total &3,223 &799,408 &2,057 &516,884  \\
    \hline
\end{tabular}
}
\end{table}

\subsection{Generation and Imputation}

\begin{figure}
    \centering
    \includegraphics[width=\linewidth]{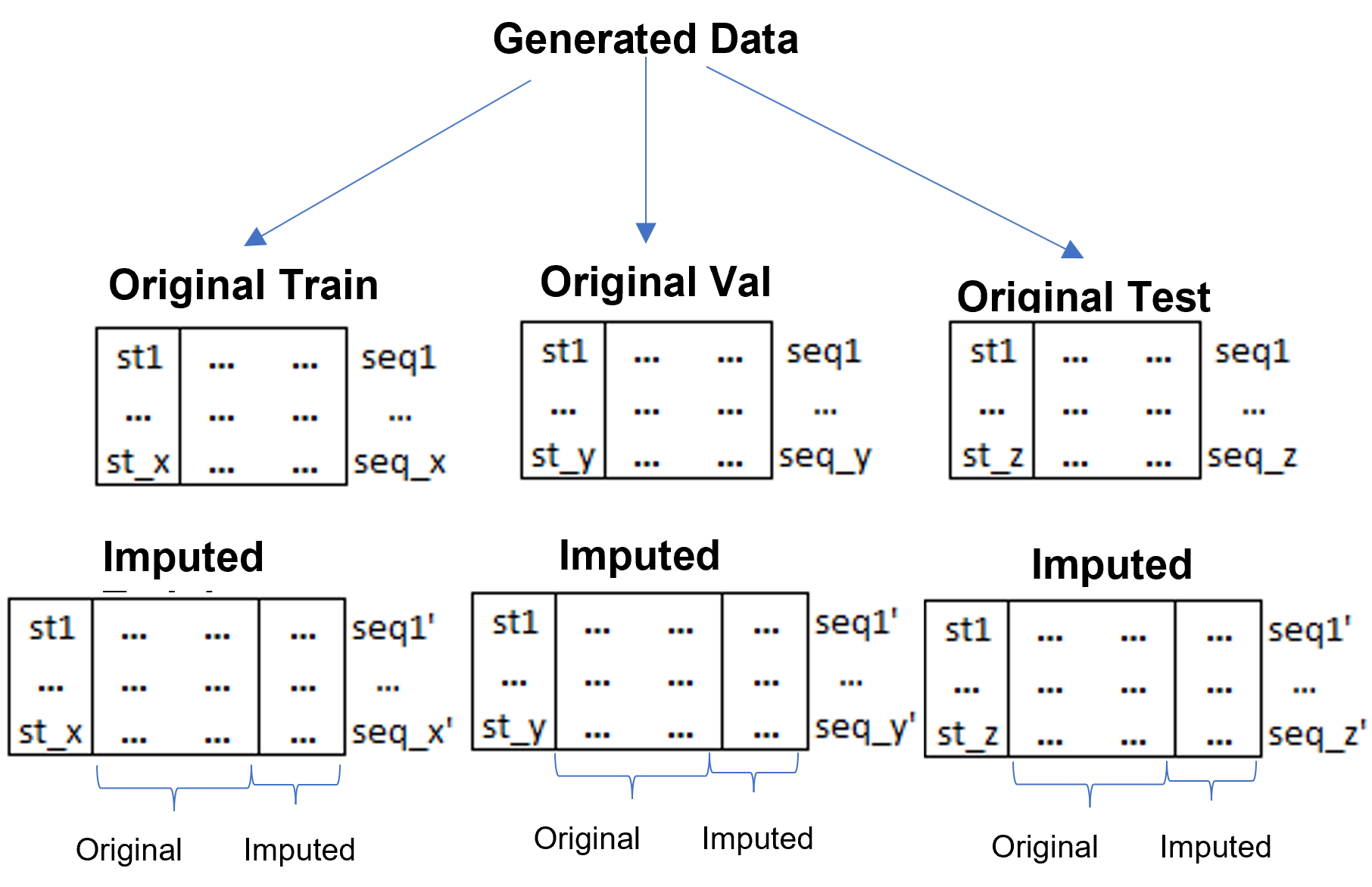}
    \caption{An Illustration of the Data Imputation Process. }
    \label{fig:generate_process}
\end{figure}

 We train three generative models: VAE-NS (non-subject), VAE 
 (subject-based) and LVAE (subject-based) to generate missing data. Since LVAE is only possible to train if we have descriptor information, which relies on student ID information, we do not apply non-subject training for LVAE. After data is generated for all the missing data, we impute back the generated data from VAE-NS by the row-number splits and impute the generated data from VAE and LVAE by ID splits (see in Figure \ref{fig:generate_process}). We do not only impute back the generated data to the train set because we believe the data augmentation on all the train, val and test sets will make the model performance harder to improve than we only augment the train set but leave the test set the same.
 
\subsection{Downstream Prediction}

There are two rounds of downstream predictions. The initial round is conducted on the original data using the three padding strategies to find the best performance model so that we can use it to predict the target variable for the generate data. The second round is a retraining round where we impute back the generated data using the best padding strategy. The second round has two parts: (i) we conduct the retraining on the combined data that contains all the generated data and the original data by IDs (for VAE, LVAE) and by row number (for VAE-NS); (ii) we conduct retraining  on the combined data with a fraction (i.e., 10/20/30/50/80/100\%) of the generated data and the original data only by IDs (for VAE, LVAE) because VAE-NS does not show salient improvement with the data it generates.

\begin{table}
  \caption{Average RMSE by Padding Strategy, Models and Data sets. The boldface represents the best performance.}
  \label{tab:na_fill_table}
  \resizebox{\columnwidth}{!}{
  \begin{tabular}{|c|c|c|c|c|c|c|}
    \hline
     \multirow{2}{*}{Avg. RMSE}  & 
    \multicolumn{3}{c|}{Geometry (Geom) }&\multicolumn{3}{c|}{Algebra II (Alg2)} \\ 
    \cline{2-7}
    & adaRNN & LSTM& Transformer &adaRNN&LSTM& Transformer\\
    \hline
    Bfill  & 0.50734 &0.47665 &0.48613& 0.52034 &\textbf{0.48967} &0.49463\\
 
    Ffill&0.51946 & \textbf{0.47664}& 0.49713& 0.51895 &0.48995 &0.49632\\
   
    Zero &\textbf{0.48160}& 0.47702& \textbf{0.40208}&\textbf{0.48860} & 0.49173 &\textbf{0.45138} \\
    
    \hline
\end{tabular}
}
\end{table}

\section{Results}\label{results}

\subsection{Evaluating the Quality of the Generated Data}

\begin{figure}[!h]
    \centering

   \subfloat[Geom Data Set]{\includegraphics[width=\linewidth]{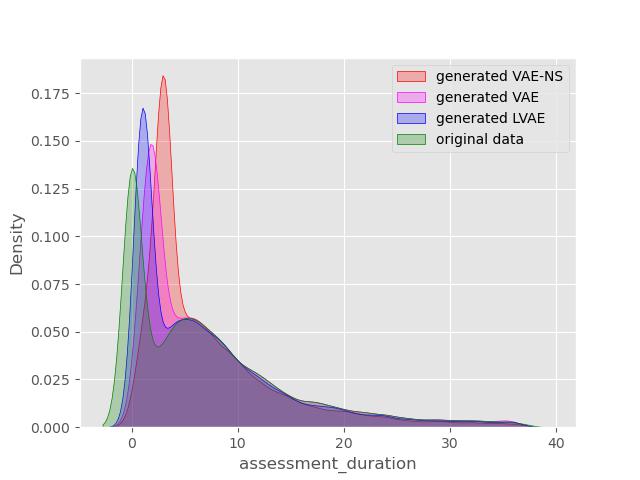}}
 
    \subfloat[Alg2 Data Set]{ \includegraphics[width=\linewidth]{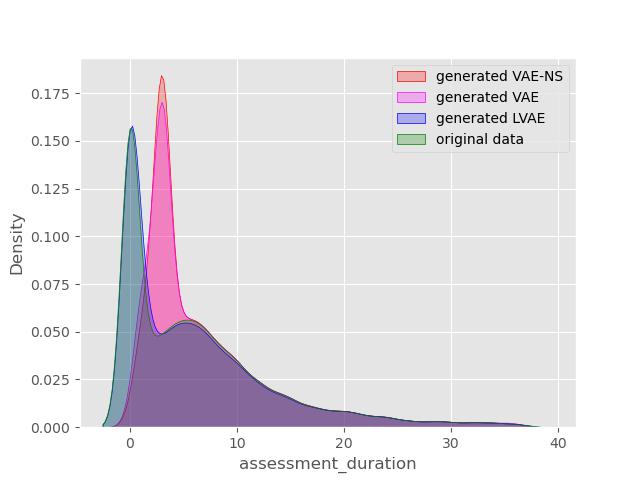}}

    \caption{`Assessment Duration' Feature Distribution Comparison Between Original Data and Generated Data}
    \label{fig:impute_qual}
\end{figure}

We exhaustively train three generative models (i.e., VAE-NS, VAE and LVAE) until its loss stops improving with different sets of hyper-parameter tuning to reach the best result. We observe that VAE-NS model is hard to converge and stops early with final loss of around 13.4349 for Alg2 data set and 0.6282 for geom data set. VAE is able to decrease loss to 0.2652 for Alg2 data set and 0.2661 for geom data set. LVAE however can decrease loss to 0.2291 for alg2 data and 0.1863 for geom data set with the latent dimension as 64 and hidden dimension as 128. Figure \ref{fig:impute_qual} selects the `assessment\_duration' feature to compare the data distribution between original data and generated data by VAE-NS, VAE and LVAE. We can tell that the Geom generated data for the feature `assessment\_duration' from VAE-NS sways the farthest from the original data whereas the generated data from VAE and LVAE are closer to the original data distribution with LVAE slightly better. The same case applies to the Alg2 data. The plot also shows that generally both VAE and LVAE can reconstruct data closely to the original data distribution, indicating that we are safe to use such generated data for downstream prediction tasks.

\subsection{Evaluating the Effectiveness of the Imputed Data}

Before we impute the generated data, we conduct the first round of downstream prediction via three models (i.e., LSTM, adaRNN and Transformer) by three padding strategies (i.e.,  bfill, ffill and zero padding) to select the best model performance as the baseline original data model performance. We run 10 random seeds for each model, padding strategy and data set. Table \ref{tab:na_fill_table} shows the detailed average RMSE for each model by padding schemes. We see that adaRNN and Transformer model obtain the best performances when padded with zero whereas LSTM model obtains its best performance via ffill padding for Geom data and bfill for Alg2 data. We then use the best performing model to predict target variable for the generated data and impute back the generated data (with the target variable) to the original data set for retraining. 

\begin{table}
  \caption{Average RMSE by Generative Models for Prediction Tasks. The bold face represents the best performance.}
  \label{tab:model_improved}
  \resizebox{\columnwidth}{!}{
  \begin{tabular}{|c|c|c|c|c|c|c|}
    \hline
     \multirow{2}{*}{Avg. RMSE} & 
    \multicolumn{3}{c|}{Geometry (Geom) }&\multicolumn{3}{c|}{Algebra II (Alg2)} \\ 
    \cline{2-7}
     & adaRNN & LSTM& Transformer &adaRNN&LSTM& Transformer\\
    \hline

    Original &0.48160& 0.47664& 0.40208&0.48860 & 0.49173 &0.45138 \\
    VAE-NS &0.58251 &0.48030& 0.37090&0.49388&0.48989&0.34071\\
    VAE&0.26570&0.26559&0.32902&0.30304&\textbf{0.27293}&\textbf{0.35260}\\
    LVAE& \textbf{0.26226}&\textbf{0.26185}&\textbf{0.28913}&\textbf{0.29470}&0.27326&0.35911\\

    \hline
\end{tabular}
}
\end{table}

\begin{figure}
    \centering

    \subfloat[Geom Data Set]{\includegraphics[width=\columnwidth]{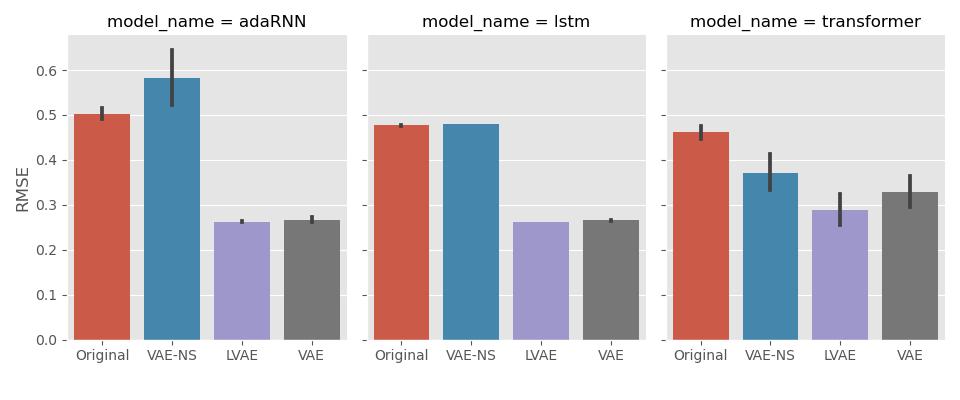}}

    \subfloat[Alg2 Data Set]{\includegraphics[width=\columnwidth]{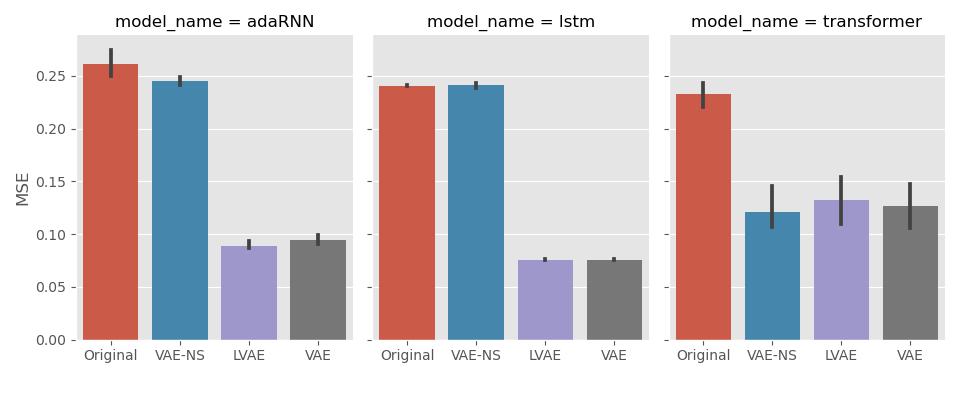}}
 
    \caption{Average Retraining RMSE Using Generated Data From Different Models. The error bar shows the min. and max. of the 10 random seeds.}
    \label{fig:ns_vs_s}
\end{figure}

We observe the retrained model performance surpasses the original model performance by big margins (see in Table \ref{tab:model_improved}). From the table on column 1 under Geom data, we observe the retrained adaRNN model performance using the generated data from VAE is 0.26570, almost about 50\% lower than the original model performance of 0.48160 in RMSE. Oppositely, the retrained model performance using the generated data from VAE-NS has RMSE of 0.58251, which is higher than the original model RMSE. This might indicate the generated data from non-subject based training perturbs the original data and creates negative gain. Further, we notice the model performance of using generated data from LVAE is even slightly better than VAE with a lower average RMSE of 0.26226. This phenomenon is present across the three models for Geom data. For Alg2 data, we also observe superior performance from both VAE and LVAE. However, the retrained model using generated data from VAE seems to perform slightly better than the one with LVAE generated data. Figure \ref{fig:ns_vs_s} visually presents the sharp drop of the average RMSE after imputing the generated data from both VAE and LVAE models.

\subsection{Evaluating the Robustness of the Imputed Data}
\begin{figure}
    \centering

    \subfloat[Geom Data Set]{\includegraphics[width=\columnwidth]{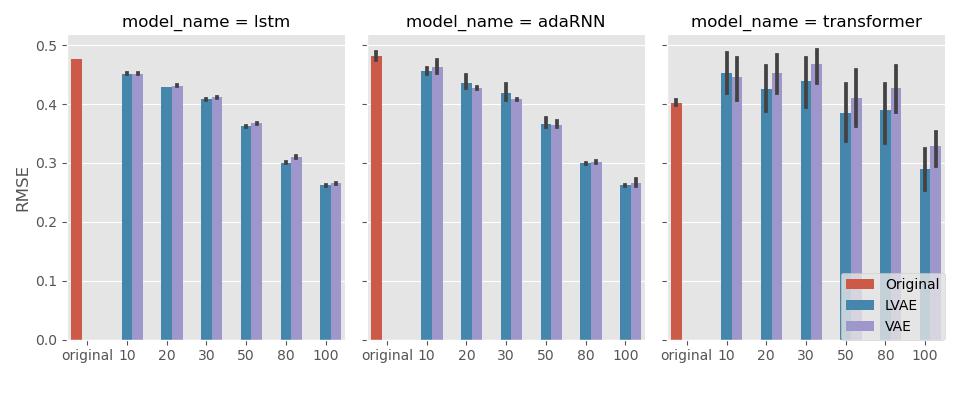}}

    \subfloat[Alg2 Data Set]{\includegraphics[width=\columnwidth]{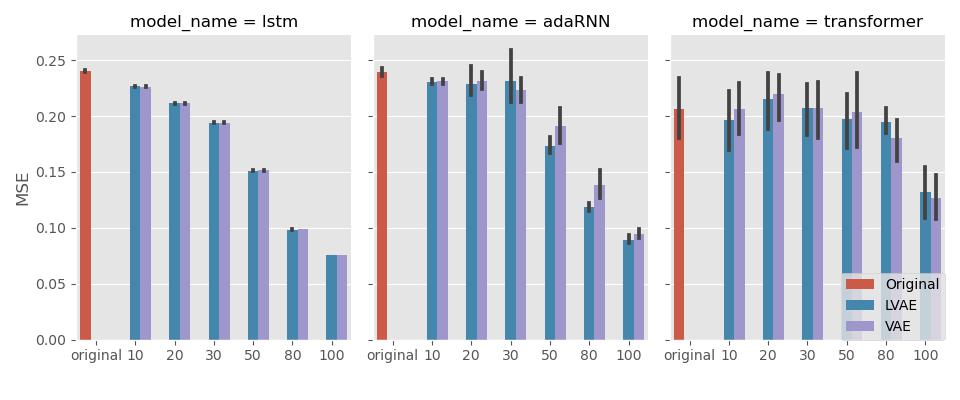}}

    \caption{Average RMSE by \% of Imputed Data vs. Original Data.}
    \label{fig:perc_data}
\end{figure}

Once we learn that imputed data can boost original model performance to a significant extent, we further experiment to validate the robustness of the imputed data. More specifically, we impute the number of students in the fraction of 10\%, 20\%, 30\%, 50\%, 80\%, 100\% back to their original train/val/test sets. The choice of percentage increments in number of students are arbitrary but all the students are linked back via their IDs to the original train/val/test sets. It is designed this way so that it is harder for the retrained model to outperform the original model as the number of students in the train/val/test set are still the same but with longer sequences. Figure \ref{fig:perc_data} showcases the effectiveness of adding different fractions of students to boost the original model performance. For LSTM and adaRNN model, we observe that the model performance starts to boost after imputing only 10\% of student IDs back. As the percentage gets higher, we see higher boosting. For Transformer model, it starts to boost after imputing 50\% of student IDs back. This confirms a known fact that large models such as Transformer model needs more data to boost its performance. In general, imputing data based on the subjects can boost the model to a great extent. 

\section{Summary}

In conclusion, to augment missing data in KT field, we first identified missing values by school testing schedules and then we train two deep generative models (i.e., VAE and LVAE) to generate quality data in the subject-based setting for imputation. With the imputed data, we are able to boost the original model by almost 50\% in average RMSE. In addition, we validate the robustness of the imputed data and observe that only 10\% of students data are needed to boost the original model performance for small to medium models such as LSTM and adaRNN and 50\% of students data are needed to boost large models such as Transformer. In future, we plan to test the effectiveness of training using the varying length, instead of fixed length, on the model performance. 

\section{Acknowledgments}
The work was mainly supported by NSF awards (1940076). Meanwhile, K12/Stride does not release student-level data to external research, but does use such data to continually improve the learning experience for students served in K12-powered schools. The authors are grateful to K12 for allowing its internal researchers to align research questions of general academic interest and publish some of those learnings.
\bibliography{references.bib}

\end{document}